\documentclass[times, review, 10pt]{elsarticle}
\usepackage[
  a4paper,
  top=4.3cm,
  right=4.8cm,
  bottom=4.3cm,
  left=4.8cm
]{geometry}

\usepackage{fontspec}
\defaultfontfeatures{Ligatures=TeX}
\IfFontExistsTF{Times New Roman}{
  \setmainfont{Times New Roman}
}{
  \setmainfont{texgyretermes}[
    Extension = .otf,
    UprightFont = *-regular,
    BoldFont = *-bold,
    ItalicFont = *-italic,
    BoldItalicFont = *-bolditalic
  ]
}
\usepackage{amssymb}
\usepackage{amsmath}
\usepackage{bbding}

\usepackage{multirow}
\usepackage{makecell}
\usepackage{booktabs}
\usepackage{float}
\usepackage{algorithm}
\usepackage{algorithmic}

\usepackage{balance}
\usepackage{array,threeparttable}
\usepackage{bm}
\usepackage{tabularx}
\usepackage{url}
\usepackage[table]{xcolor}
\definecolor{oursrow}{HTML}{EAF5FF}
\usepackage[colorlinks=true,linkcolor=blue,citecolor=blue,urlcolor=blue]{hyperref}

\usepackage{caption}
\begin{document}
\begin{frontmatter}
  \title{TAQ-MOTR: Tracking-Aware Detection Query Formation for End-to-End Multiple Object Tracking}
  \cortext[cor1]{Corresponding author}
  \affiliation[a_1]{organization={School of Astronautics, Beihang University},
    addressline={37 Xueyuan Road, Haidian District},
    city={Beijing 100191},
    country={China}
  }
  \affiliation[a_2]{organization={School of Software, Beihang University},
    addressline={37 Xueyuan Road, Haidian District},
    city={Beijing 100191},
    country={China}}

  \affiliation[a_3]{op={},
    organization={Key Laboratory of Spacecraft Design Optimization and Dynamic Simulation Technology, Ministry of Education}}
  \affiliation[a_4]{op={},
    organization={State Key Laboratory of High-Efficiency Reusable Aerospace Transportation Technology}}

  \author[a_1]{Hanyang Liu}
  \ead{hanyangliu@buaa.edu.cn}

  \author[a_1,a_3]{Hong Zhang}
  \ead{dmrzhang@buaa.edu.cn}

  \author[a_1,a_4]{Ding Yuan}
  \ead{dyuan@buaa.edu.cn}

  \author[a_2]{Yang Han}
  \ead{hanyang2022@buaa.edu.cn}

  \author[a_1,a_4]{Yifan Yang\corref{cor1}}
  \ead{stephenyoung@buaa.edu.cn}

  \begin{abstract}
    Tracking-by-propagation (TBP) methods perform end-to-end multi-object tracking by propagating track queries across frames, jointly handling detection and identity association without heuristic post-processing while naturally modeling temporal information.
    However, as detection and association are tightly coupled, detection queries in TBP are expected to discover currently untracked objects and provide complementary current-frame evidence for challenging tracks whose queries and identities are retained.
    Existing methods largely inherit detection-query formation from single-frame detection and construct such queries before the decoder with little awareness of trajectory states.
    Consequently, currently untracked objects may be missed, while challenging tracked objects with reduced visibility under occlusion may lack complementary evidence, causing association failures or duplicate responses.
    To address these issues, we propose TAQ-MOTR, a tracking-aware detection query formation framework that employs a pre-decoder Hybrid Query Generator (HQG) to fuse current-frame image features, track memory, and learnable memory. By conditioning detection queries on trajectory states, TAQ-MOTR adapts them to the discovery of currently untracked objects and tracked-object complementation.
    We further introduce a training-only auxiliary detection branch whose one-to-one label assignment is decoupled from that of the main tracking branch, providing additional detection supervision to strengthen the shared HQG and decoder for the discovery of currently untracked objects.
    TAQ-MOTR improves performance on DanceTrack, SportsMOT, and BFT, achieving 72.0 HOTA, 63.1 AssA, and 76.6 IDF1 on the DanceTrack test set.
    Code is available at \url{https://github.com/Magicmiya/TAQ-MOTR}.
\end{abstract}


\end{frontmatter}

\section{Introduction}

Multi-object tracking (MOT) aims to localize objects of interest in a video and preserve their identities across frames. It supports applications in autonomous driving, intelligent transportation, video surveillance~\cite{ciaparrone2020deep,luo2021multiple}, and sports analytics~\cite{cui2023sportsmot}. A central challenge in MOT is to jointly handle object discovery and identity preservation under complex motion, occlusion, and appearance variation. Traditional methods usually follow the Tracking-by-Detection (TBD) paradigm, which first detects objects frame by frame and then associates detections over time. Although representative methods such as ByteTrack~\cite{zhang2022bytetrack} are effective, their detectors are optimized mainly for frame-level recognition and are not explicitly conditioned on trajectory states, making detection errors directly affect association.

Inspired by DETR-style detectors~\cite{zhu2021deformable}, recent end-to-end MOT methods extend query-based set prediction to videos by propagating identity-preserving track queries across frames~\cite{sun2020transtrack,meinhardt2022trackformer,zeng2022motr}. We refer to this family as Tracking-by-Propagation (TBP). Unlike DETR object queries, which detect all objects in a single image, TBP detection queries operate alongside propagated track queries that maintain existing trajectories. They therefore serve two roles: discovering currently untracked objects and supplying complementary current-frame evidence to challenging retained tracks. Here, currently untracked objects are those without a maintained track query, including newly appearing objects and previously observed objects that have not been initialized or whose tracks have been terminated. Temporarily lost objects with retained track queries remain tracked, even if no tracking result is output in the current frame. Under identity-preserving one-to-one assignment, ground-truth objects associated with propagated track queries are reserved for those queries, leaving detection queries with positive labels only from currently untracked objects. This imbalanced label assignment leaves the detection component insufficiently trained, contributing to the conflict between detection and association~\cite{yu2023motrv3}.

\begin{figure}[htbp!]
  \centering
  \includegraphics[width=\linewidth]{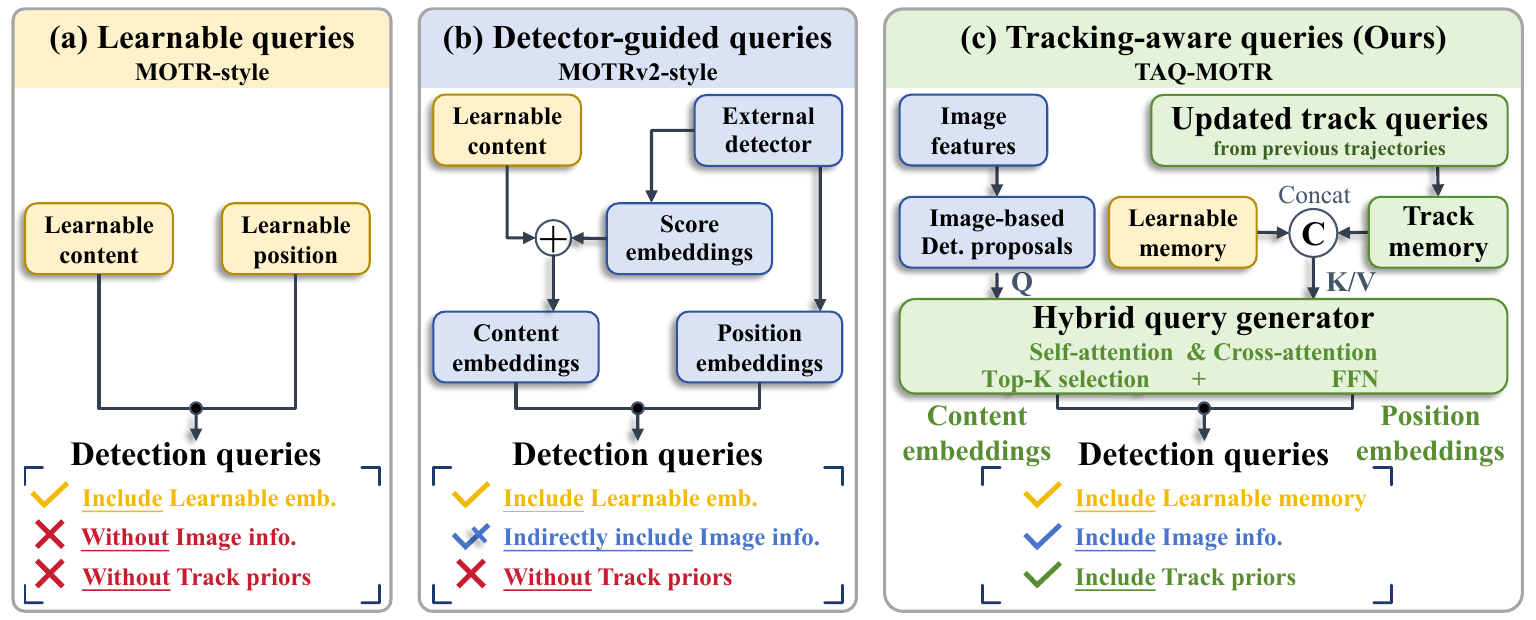}
  \caption{\textbf{Detection query formation before decoding.}
    (a) MOTR: learnable queries; (b) MOTRv2: detector-guided queries; (c) TAQ-MOTR: tracking-aware queries.
    The comparison highlights current-frame information, trajectory-state priors, and learnable embeddings.}
  \label{fig:query_paradigms}
\end{figure}

Existing TBP methods use different query-initialization strategies to support object discovery. MOTR~\cite{zeng2022motr} initializes detection queries with learnable content and positional embeddings (Fig.~\ref{fig:query_paradigms}(a)). Building on learnable query embeddings, MOTRv2~\cite{zhang2023motrv2} uses boxes and confidence scores from a pretrained external detector to initialize detection queries (Fig.~\ref{fig:query_paradigms}(b)). This introduces current-frame information indirectly and reduces the tracker's detection burden, but requires an additional detector. Neither strategy conditions detection query formation on existing trajectory states; distinguishing tracked from currently untracked objects still relies on subsequent interaction with track queries in the decoder.

Motivated by visual search studies showing that human visual attention is guided by task demands toward informative regions~\cite{hayhoe2005eye,wolfe1994guided}, we seek to align detection query formation with current tracking needs. We therefore propose TAQ-MOTR, a tracking-aware detection query formation framework for end-to-end MOT (Fig.~\ref{fig:query_paradigms}(c)). The Hybrid Query Generator (HQG) makes detection-query initialization tracking-aware by incorporating existing trajectory states before shared decoding.
Specifically, HQG refines current-frame image-based proposals through attention to track memory and learnable memory, which provide track-state information and trajectory-independent detection priors, respectively.
The conditioned proposals are then scored and selected by Top-K to supply the detection-query content embeddings and reference boxes, incorporating tracking context before their interaction with track queries in the decoder.

Furthermore, we add a training-only auxiliary detection branch to improve detection-side learning under sparse supervision. It reuses the image-based detection proposals and shares HQG and decoder parameters with the main branch, but excludes track memory and blocks interaction with the main queries through attention masks. Its proposal and decoder predictions are independently matched to all current-frame ground-truth objects, preserving one-to-one assignment within each branch and supervising HQG and the decoder through their respective losses. This design supports joint optimization within the end-to-end tracker without an external detector or additional trainable parameters, and the auxiliary branch is removed at inference.

We evaluate TAQ-MOTR on DanceTrack~\cite{sun2022dancetrack}, SportsMOT~\cite{cui2023sportsmot}, and BFT~\cite{zheng2024nettrack}, achieving the highest HOTA among the compared TBP methods on all three benchmarks. The association gains and competitive detection performance support the benefit of forming detection queries with awareness of existing track states.

The main contributions of this work are summarized as follows:
\begin{itemize}
  \item We propose TAQ-MOTR to make detection query formation tracking-aware in TBP-based MOT. It introduces trajectory-state conditioning before decoder interaction, enabling detection queries to adapt to their shifted role in query propagation while retaining the unified TBP decoding framework.

  \item We design a Hybrid Query Generator (HQG) for tracking-aware detection-query initialization.
By conditioning image-based proposals on track memory and learnable memory before scoring and Top-K selection, HQG incorporates existing trajectory states into detection queries before shared decoding.

  \item We introduce a training-only auxiliary detection branch with a stage-wise training strategy to provide additional detection supervision through decoupled label assignments between the auxiliary and main tracking branches. By strengthening detection query learning without competing with track queries for labels, this branch improves detection quality and further benefits identity association.
\end{itemize}

\section{Related Work}

\subsection{Multi-object Tracking Paradigms}

Multi-object tracking (MOT) methods can be broadly categorized by how object detection and temporal association are coupled. Tracking-by-Detection (TBD) first performs frame-wise detection and then associates detection results across frames. Representative trackers, such as ByteTrack~\cite{zhang2022bytetrack}, OC-SORT~\cite{cao2022ocsort} and Hybrid-SORT~\cite{yang2024hybridsort}, improve this pipeline by recovering low-confidence detections, refining motion modeling, or combining complementary association cues. Despite their effectiveness, the separated pipeline makes track initialization and identity continuity rely heavily on detector quality and hand-crafted matching. Missed or inaccurate detections may therefore lead to target loss, trajectory fragmentation, and identity switches.

Inspired by DETR~\cite{carion2020end}, recent end-to-end MOT methods couple detection and association through query-based set prediction. TransTrack~\cite{sun2020transtrack} uses target features from the previous frame as tracking queries, while TrackFormer~\cite{meinhardt2022trackformer} and MOTR~\cite{zeng2022motr} further develop query propagation for identity-preserving tracking. MeMOTR~\cite{gao2023memotr} further incorporates long-term memory into track-query updates to maintain identity consistency, while FGO-PMB~\cite{jin2026fgopmb} uses multi-frame factor graph optimization to refine trajectory states in 3D MOT. In the TBP paradigm, propagated track queries and detection queries are jointly decoded in a shared decoder, enabling unified optimization of object discovery and identity association. However, this tight coupling also changes the role of detection queries. Beyond the discovery of currently untracked objects, detection queries are expected to provide complementary current-frame evidence for challenging tracked objects under occlusion, crossing, or overlap. This dual role makes detection query formation before shared decoding crucial for TBP-based MOT.

Another line of end-to-end MOT revisits the separation between object detection and identity association. LA-MOTR~\cite{wang2025motr} decouples object detection and trajectory association into dedicated modules, while MOTIP~\cite{gao2025multiple} formulates association as in-context ID prediction over frame-level object representations. Although these methods reduce the entanglement between detection and association, their detection stage is largely driven by single-frame object evidence before identity reasoning. As a result, detection is less directly conditioned on propagated track states, leaving limited room for track memory to guide current-frame object discovery and complement challenging tracked objects. TAQ-MOTR therefore focuses on a different setting: it preserves the TBP formulation with a unified query-propagation decoder, and improves detection query formation before decoding by making detection queries tracking-aware.

\subsection{Detection Query Formation in End-to-End MOT}
\label{sec:rw_query_formation}

End-to-end MOT methods such as MOTR~\cite{zeng2022motr} build on DETR-style object detectors~\cite{carion2020end}. In MOTR, detection queries are initialized with a fixed-size set of learnable content and positional embeddings. These embeddings acquire shared detection priors from the training data, providing a consistent initialization across frames. However, they are not conditioned on current-frame observations or existing track states. Detection queries therefore rely on subsequent interactions with image features and track queries in the decoder to localize objects and distinguish currently untracked objects from those already tracked. This lack of guidance at initialization can leave early decoder attention insufficiently focused on relevant object regions, limiting how effectively detection queries extract features for tracking. To incorporate current-frame object information, MOTRv2~\cite{zhang2023motrv2} uses proposals from a pretrained external detector to initialize detection queries, but its proposal generation does not incorporate existing track states or undergo joint optimization with the tracker.

Contextual conditioning in related vision tasks provides further inspiration for query formation. SAM~2~\cite{ravi2025sam} conditions current-frame image representations on object-specific memory for segmentation, with memory attention and mask decoding performed separately for each object. SAM2MOT~\cite{jiang2026sam2mot}, for example, adds a separate cross-object interaction module to coordinate memory updates under occlusion while retaining per-object segmentation. In object detection, RT-DETRv2~\cite{lv2024rt} selects scored image-based proposals to initialize decoder queries and jointly trains proposal generation with all-object supervision. Extending image-based proposal initialization to TBP also requires conditioning on existing trajectory states.

HQG extends image-based proposal initialization with track-state conditioning before joint MOT decoding, while preserving the unified TBP framework.

\subsection{Query Competition and Detection-side Supervision}

Supervision strategies in end-to-end MOT differ in how they balance detection learning with identity-preserving tracking.

MOTRv3~\cite{yu2023motrv3} redistributes labels between detection and track queries through release-fetch supervision, while CO-MOT~\cite{luo2025co} extends detection supervision to tracked objects in intermediate decoder layers and introduces shadow query sets. These strategies modify supervision within the main decoder. Our approach instead uses a separate auxiliary query group with independent all-object matching, drawing on query grouping and attention masking in DN-DETR~\cite{li2022dn}. This supplies additional detection supervision while preserving the main branch's identity-preserving assignments.

\section{Our Approach}

\subsection{Overall Architecture}
\begin{figure}[htbp!]
    \centering
    \includegraphics[width=\linewidth]{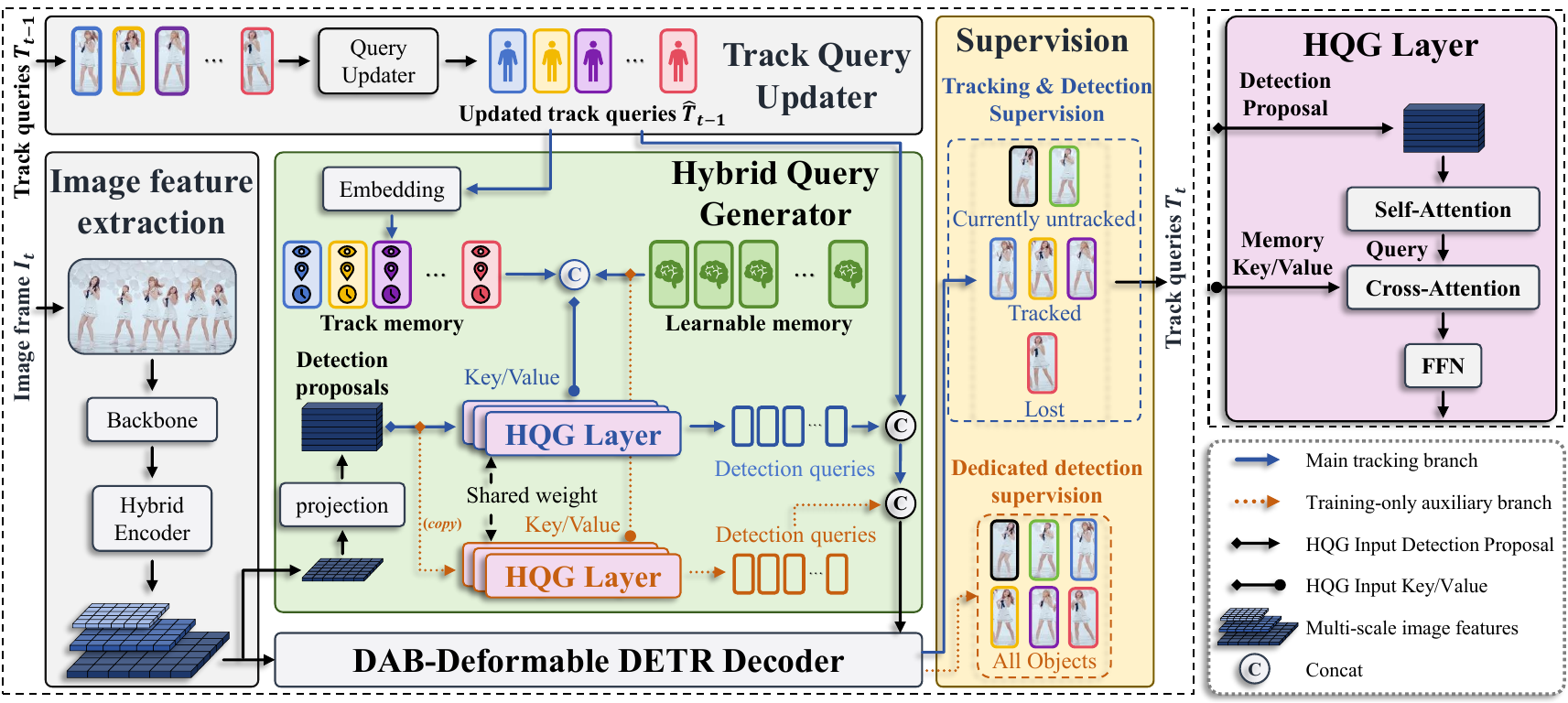}
    \caption{\textbf{Overall architecture of TAQ-MOTR.}
        TAQ-MOTR forms tracking-aware detection queries before decoding. In the main tracking branch, HQG uses image-based detection proposals as queries and reads track memory and learnable memory as keys/values. The generated detection queries are concatenated with updated track queries and jointly decoded. During training, an auxiliary detection branch shares the detection proposals and HQG layers but removes track memory, producing a separate detection-query group supervised by all current-frame objects. Attention masks isolate this auxiliary group from the main query groups. The right panel shows an HQG layer.}
    \label{fig:overall_architecture}
\end{figure}

TAQ-MOTR is an end-to-end MOT framework that forms tracking-aware detection queries before decoder interaction. As shown in Fig.~\ref{fig:overall_architecture}, given an input frame $I_t$, a backbone and an RT-DETR-style hybrid encoder extract multi-scale image features~\cite{zhao2024detrs}. These features serve as visual context for the DAB-Deformable DETR decoder~\cite{liu2022dabdetr,zhu2021deformable}, while the lowest-resolution feature map is projected into image-based detection proposals for the Hybrid Query Generator (HQG). In parallel, track queries $T_{t-1}$ propagated from the previous frame are updated by the track-query updater. The updated propagated track queries are used as decoder queries for identity-preserving tracking and are also converted into track memory for HQG.

The main tracking branch is active during both training and inference. HQG takes the detection proposals as input queries and allows them to interact with two memory sources: track memory, which encodes existing trajectory states, and learnable memory, which stores trajectory-agnostic detection priors. The resulting detection queries therefore combine current-frame object evidence, track-state priors, and learned detection priors before entering the decoder. They are concatenated with the updated track queries and jointly decoded. High-confidence detection-query outputs are initialized as new track queries, while active propagated track queries are updated to continue existing trajectories. Following the missing-frame tolerance principle in MOTR~\cite{zeng2022motr} and MeMOTR~\cite{gao2023memotr}, established tracks are retained during temporary loss until the configured missing-frame tolerance is reached. Retained track queries continue to participate in decoding; once a retained query again produces a reliable prediction, tracking resumes with the same identity. These objects remain tracked even when no current-frame output box is produced. Exclusion from HQG's conditioning memory does not itself delete a retained track query or its identity. Once the tolerance is reached, the track query is deleted, and its ID is permanently retired and will not be restored or reassigned. A later detection of the same physical object therefore initializes a new track with a new identity. The newly initialized and retained track queries together form $T_t$ for propagation to the next frame.

For tracked-object complementation, HQG conditions detection queries on current-frame image evidence, track memory, and learnable memory. Following MOTR~\cite{zeng2022motr}, retained track queries read detection-query features through decoder self-attention to complement the representations used for their predictions, while the existing identities remain assigned to the track queries.

The training-only auxiliary branch strengthens detection-side learning without changing the inference pipeline. It reuses the same detection proposals and HQG-layer weights as the main branch, but excludes track memory and reads only learnable memory. The auxiliary queries are decoded together with the main queries, isolated by attention masks, and supervised by all ground-truth objects in the current frame. They are not used for tracking prediction and are not propagated to subsequent frames.

\subsection{Hybrid Query Generator}
\label{sec:hqg}
\begin{figure}[t]
    \centering
    \includegraphics[width=\linewidth]{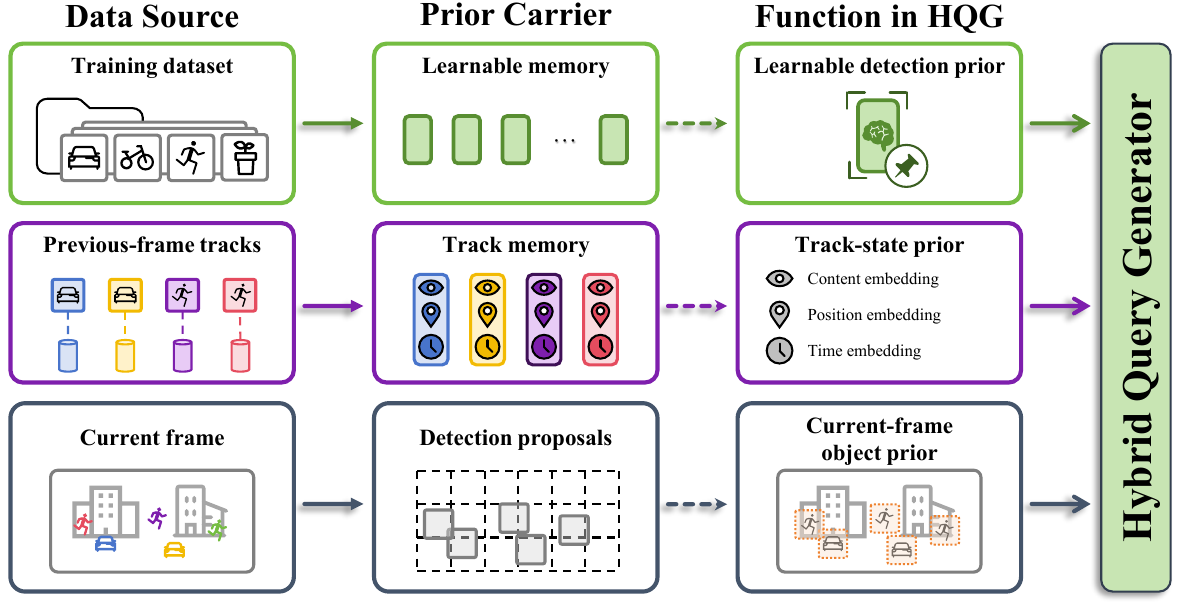}
    \caption{\textbf{Prior carriers used in HQG.}
        Learnable memory provides learnable detection priors that are independent of specific trajectories. Track memory encodes existing trajectory states from propagated track queries, including content, position, and disappearance-time information. Detection proposals provide current-frame object evidence and coarse spatial priors.}
    \label{fig:hqg_priors}
\end{figure}

As discussed in Section~\ref{sec:rw_query_formation}, detection queries in TBP-based MOT do not simply inherit DETR's full-image discovery role. Existing targets are mainly maintained by propagated track queries, whereas detection queries should discover currently untracked objects and provide complementary current-frame evidence for tracked targets under occlusion, crossing, or overlap. HQG addresses this tracking-specific role by organizing three prior carriers before decoding: learnable memory, track memory, and image-based detection proposals. Its position in the overall architecture is shown in Fig.~\ref{fig:overall_architecture}, and the three prior carriers are decomposed row by row in Fig.~\ref{fig:hqg_priors}.

\paragraph{Learnable memory}
As shown in the top row of Fig.~\ref{fig:hqg_priors}, learnable memory is a single set of trainable tokens shared across all HQG layers and the main and auxiliary branches. It learns trajectory-independent detection priors from the training dataset, preserving the prior role of learnable queries in DETR/MOTR-style models. The tokens provide cross-attention keys and values alongside track memory, with image-based detection proposals as queries. They remain available when track memory is empty and are reused across frames, but are neither directly decoded nor propagated as track states. We denote this shared memory as
\begin{equation}
    \mathbf{M}^{\mathrm{learn}} \in \mathbb{R}^{N_l \times d},
    \label{eq:learnable_memory}
\end{equation}
where $N_l$ is the number of tokens in the shared memory set and $d$ is the embedding dimension. The same $\mathbf{M}^{\mathrm{learn}}$ is used in every HQG layer.

\paragraph{Track memory}
As shown in the middle row of Fig.~\ref{fig:hqg_priors}, track memory carries track-state priors from previous-frame trajectories. In the overall architecture of Fig.~\ref{fig:overall_architecture}, it is derived from the updated propagated track queries and then used as part of the HQG key/value memory. Each track-memory token contains a content prior from the track-query feature, a spatial prior from the reference box, and a temporal-state prior from the disappearance-time index. Let $C_t^{\mathrm{trk}}$, $B_t^{\mathrm{trk}}$, and $\Delta_t^{\mathrm{trk}}$ denote the track content embedding, reference box, and disappearance-time index, respectively. Track memory is formulated as
\begin{equation}
    \mathbf{M}_t^{\mathrm{trk}} = C_t^{\mathrm{trk}} + \mathrm{FoPE}(B_t^{\mathrm{trk}}) + \mathbf{E}^{\mathrm{time}}[\Delta_t^{\mathrm{trk}}],
    \label{eq:track_memory}
\end{equation}
where $\mathrm{FoPE}(\cdot)$ is Fourier position embedding and $\mathbf{E}^{\mathrm{time}}$ is a learnable temporal embedding table. Through $\mathbf{M}_t^{\mathrm{trk}}$, detection proposals can condition query formation on existing trajectory states.

\paragraph{Detection proposals}
As shown in the bottom row of Fig.~\ref{fig:hqg_priors}, detection proposals provide current-frame object evidence and coarse spatial priors. In Fig.~\ref{fig:overall_architecture}, they are generated from the lowest-resolution encoder feature map, which provides strong semantic cues with a low token resolution. The feature map is flattened and projected into proposal content embeddings $F_t^{\mathrm{low}}$. Regular grid anchors $A_t^0$ are generated on the same feature map and encoded by Fourier position embedding. The initial detection proposals are defined as
\begin{equation}
    \widehat{\mathbf{D}}_t^{\mathrm{det}} = F_t^{\mathrm{low}} + \mathrm{FoPE}(A_t^0)\in \mathbb{R}^{N_p \times d},
    \label{eq:det_proposal}
\end{equation}
where $N_p=H_{\mathrm{low}}W_{\mathrm{low}}$, $H_{\mathrm{low}}$ and $W_{\mathrm{low}}$ are the height and width of the lowest-resolution encoder feature map, and $d$ is the embedding dimension. We define $A_t^0\in(0,1)^{N_p\times4}$ as grid anchors in normalized $(c_x,c_y,w,h)$ coordinates; $\mathrm{FoPE}$ maps them to $N_p\times d$ positional embeddings. $\widehat{\mathbf{D}}_t^{\mathrm{det}}$ carries current-frame object evidence and coarse spatial priors and serves as the input query of HQG.

\paragraph{Tracking-aware query formation}
In the main branch of Fig.~\ref{fig:overall_architecture}, HQG uses $\widehat{\mathbf{D}}_t^{\mathrm{det}}$ as queries and the concatenation of track memory and learnable memory as keys/values. Let $H_t^0=\widehat{\mathbf{D}}_t^{\mathrm{det}}$ and $\mathbf{M}_t^{\mathrm{hyb}}=[\mathbf{M}_t^{\mathrm{trk}},\mathbf{M}^{\mathrm{learn}}]$. Each HQG layer, illustrated in the right panel of Fig.~\ref{fig:overall_architecture}, contains self-attention, cross-attention, and an FFN:
\begin{equation}
    \begin{aligned}
        \widetilde{H}_t^l & = \operatorname{SelfAttention}(H_t^l),                                                               \\
        H_t^{l+1}         & = \mathrm{FFN}\big(\operatorname{CrossAttention}(\widetilde{H}_t^l,\mathbf{M}_t^{\mathrm{hyb}})\big),
    \end{aligned}
    \label{eq:hqg_block}
\end{equation}
where $\mathbf{M}_t^{\mathrm{hyb}}$ provides the cross-attention keys and values. Residual connections and normalization are omitted for clarity. After $L$ HQG layers, the output is projected and scored:
\begin{equation}
    \begin{aligned}
        E_t                & =        \mathrm{Proj}(H_t^L),                                \\
        S_t                & =        \psi_{\mathrm{cls}}(E_t),                            \\
        B_t                & =        \sigma(\psi_{\mathrm{box}}(E_t)+\sigma^{-1}(A_t^0)), \\
        Q_t^{\mathrm{det}} & =        \mathrm{TopK}(E_t,S_t,B_t).
    \end{aligned}
    \label{eq:hqg_topk}
\end{equation}
Here, $E_t$ and $S_t$ denote candidate embeddings and classification scores, and $B_t$ contains normalized predicted boxes. $\sigma$ is the sigmoid function, and $\psi_{\mathrm{box}}$ predicts residuals in inverse-sigmoid coordinates. $\mathrm{TopK}$ ranks candidates by $S_t$ and returns $K$ pairs of content embeddings and corresponding reference boxes, of shapes $K\times d$ and $K\times4$, respectively.
All candidates share this ranking, with no separate quotas for tracked and currently untracked objects.
Conditioning before selection makes these queries tracking-aware.

Following RT-DETRv2~\cite{lv2024rt}, only the selected $K$ candidates participate in matching and direct supervision; the remaining candidates are excluded from loss computation and decoder initialization.
The selected content embeddings and reference boxes are detached from the computation graph before decoder initialization, preventing decoder losses from backpropagating to HQG through these tensors.
Direct detection losses train HQG, whereas decoder losses train the downstream interactions; no separate complementation target is introduced.

\subsection{Detection-only Auxiliary Learning}
\label{sec:det_aux_learning}

Although HQG conditions detection queries on track states, one-to-one assignment still limits their positive supervision. Ground-truth identities held by maintained track queries, including temporarily lost queries, are reserved for those queries. Main detection queries are matched only to currently untracked objects, whose identities are not held by any maintained track query. At clip initialization, they can match all current-frame objects. Auxiliary queries are independently matched to all current-frame objects at supervised layers when their branch is active on non-initial clip frames. This supervision imbalance weakens detection-side learning, making detection queries less effective in discovering currently untracked objects and in providing complementary current-frame evidence for challenging tracked targets.

To quantify this supervision imbalance, we analyze GT assignments throughout the 30-epoch DanceTrack training run. The track-query, main detection, auxiliary detection, and combined detection ratios are $T/G$, $D/G$, $A/G$, and $(D+A)/G$, where $G$ counts all sampled GT occurrences, $T$ and $A$ are logged positive-assignment counts, and $D=G-T$ reconstructs main detection supervision. The auxiliary branch is disabled on the first frame of every training clip, where $T=A=0$ and main detection queries receive all GT labels; these frames remain in $G$.

\begin{figure}[htbp!]
    \centering
    \includegraphics[width=\linewidth]{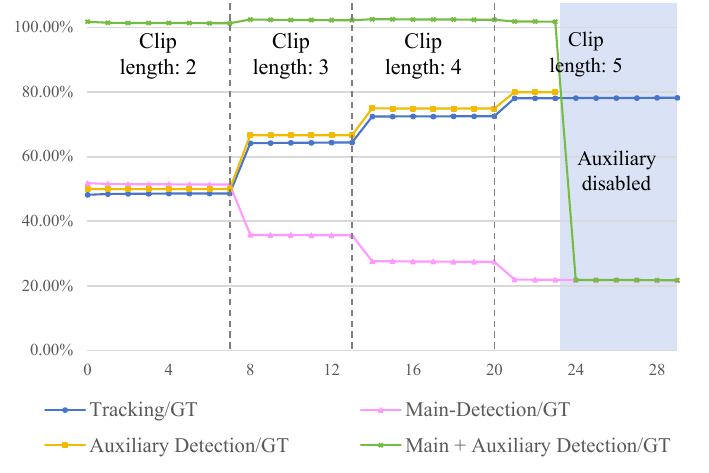}
    \caption{\textbf{Positive label assignments during training on DanceTrack.}
        All ratios use the total GT count per epoch, including clip-initial frames. The auxiliary branch is disabled on each clip's first frame, so Auxiliary Detection/GT remains below 100\% while enabled. Dashed lines mark clip-length changes; shading marks branch shutdown from epoch 24. The auxiliary-only ratio is then zero and omitted. Epoch numbering starts at 0.}
    \label{fig:label_assignments}
\end{figure}

In Fig.~\ref{fig:label_assignments}, main detection receives 51.80\% and 27.59\% of GT labels at epochs 0 and 14; excluding clip-initial frames gives 3.58\% and 3.45\%. Auxiliary Detection/GT stays below 100\% because initial frames contribute to $G$ but not $A$. On non-initial frames alone, auxiliary assignments cover 99.94\%--99.98\% of GT during epochs 0--23. Longer clips reduce the initialization-frame share, explaining the auxiliary curve's stepwise increases. The combined ratio averages 102.01\% over these epochs because labels are reused across branches, supplementing detection supervision without reallocating main-branch labels. These counts come from one run, not separately trained models.

To provide stable detection-side supervision, we introduce a training-only auxiliary detection branch, shown by the dotted auxiliary path in Fig.~\ref{fig:overall_architecture}. It follows the query grouping and attention masking strategy of DN-DETR~\cite{li2022dn}, but does not construct denoising queries from ground-truth annotations. Instead, it reuses the same image-based detection proposals as the main branch to form a separate detection-query group. The auxiliary branch shares the initial proposals $\widehat{\mathbf{D}}_t^{\mathrm{det}}$ and HQG-layer weights with the main branch, but removes track memory from cross-attention:
\begin{equation}
    \begin{aligned}
        H_{t,\mathrm{aux}}^0     & =        \widehat{\mathbf{D}}_t^{\mathrm{det}},                               \\
        H_{t,\mathrm{aux}}^{l+1} & =        \mathrm{HQGBlock}(H_{t,\mathrm{aux}}^l,\mathbf{M}^{\mathrm{learn}}).
    \end{aligned}
    \label{eq:aux_hqg}
\end{equation}
Thus, the auxiliary queries retain current-frame image evidence and learnable detection priors while remaining independent of propagated track states.

As shown in Fig.~\ref{fig:overall_architecture}, the auxiliary queries are decoded together with the main detection queries and track queries, but attention masks block information exchange between the auxiliary group and the main query groups. The main branch performs tracking prediction, track-query update, and query propagation. The auxiliary branch is used only for detection losses on non-initial clip frames when enabled by the training schedule: its outputs are independently matched with all ground-truth objects in the current frame, and its queries are discarded after training. Both branches share HQG and decoder parameters, including their prediction heads, without adding trainable parameters or competing with track queries for positive labels. This design strengthens current-frame object discovery and complementary evidence learning without increasing inference cost or disrupting the unified TBP decoder.

We follow RT-DETRv2~\cite{lv2024rt} for the loss functions.
Main-branch proposals are matched to currently untracked ground-truth objects; active auxiliary proposals are independently matched to all current-frame ground-truth objects.
Unmatched main-branch proposals, including those covering already tracked objects, are supervised as background.
The same assignment rules apply to decoder detection-query predictions, while track queries retain their identity-based assignments.\par

\section{Experiments}

\subsection{Benchmarks and Metrics}

We use three benchmarks: DanceTrack~\cite{sun2022dancetrack}, SportsMOT~\cite{cui2023sportsmot}, and Bird Flock Tracking (BFT)~\cite{zheng2024nettrack}. DanceTrack features similar appearances, crossings, and occlusions; SportsMOT introduces fast motion, viewpoint changes, and small targets in wide-view scenes. BFT extends evaluation to non-human targets with rapid motion, deformation, and occlusion. We report HOTA as the primary overall metric, DetA and AssA for detection and association quality~\cite{luiten2021hota}, IDF1 for identity preservation~\cite{ristani2016performance}, and MOTA for frame-level tracking errors~\cite{bernardin2008evaluating}. Best and second-best results are highlighted among Tracking-by-Propagation (TBP) methods, which propagate track queries across frames.

\subsection{Implementation Details}

TAQ-MOTR follows the MeMOTR framework~\cite{gao2023memotr} for query propagation and long-term memory. The model uses RT-DETRv2 weights pretrained on COCO~\cite{lv2024rt,lin2014microsoft}, a PResNet-50vd backbone, and the RT-DETRv2 Hybrid Encoder. The decoder follows DAB-Deformable DETR~\cite{liu2022dabdetr,zhu2021deformable} with six layers and takes three feature levels with strides of~8, 16, and 32. HQG constructs image-based detection proposals from the stride-32 feature level. The first decoder layer updates only detection-side queries, while the remaining layers jointly decode detection queries and propagated track queries. Unless otherwise specified, HQG uses 200 main detection queries and 200 detection-only auxiliary queries. HQG contains three layers that share a set of 32 learnable memory tokens.

Training uses NVIDIA GeForce RTX 4090 GPUs. DanceTrack is the main benchmark for selecting the default configuration and conducting ablations, and its default model is trained for 30 epochs on four GPUs. We adopt a progressive training curriculum based on MeMOTR~\cite{gao2023memotr}: training clips are sampled with random temporal intervals of up to 10 frames, and the clip length gradually increases from 2 to 5. Within this curriculum, we remove intermediate-layer auxiliary decoder losses at epoch 14 and disable the auxiliary branch at epoch 24. SportsMOT uses the same architecture with a 42-epoch schedule; the entire auxiliary branch is disabled at epoch 36, the beginning of the final training stage. BFT uses the same architecture with a 50-epoch schedule, with training stages and auxiliary-supervision transitions proportionally mapped from the DanceTrack schedule.

For tracker training in the main comparisons, TAQ-MOTR uses only the official training split of each benchmark; COCO detector pretraining is specified above and is distinct from additional tracking-training data. We additionally report a SportsMOT variant trained with additional CrowdHuman data, marked with \textsuperscript{‡} and excluded from the main ranking. TAQ-MOTR performs online tracking without offline trajectory linking, smoothing, or interpolation. Outputs undergo per-frame confidence filtering.

\subsection{Comparison with State-of-the-Art Methods}

Accuracy results for other methods follow the cited publications or official releases. We distinguish additional data for tracking-model training (\textsuperscript{‡}) from additional models (\textsuperscript{†}) and identify train+val variants explicitly. TBDQ-Net is grouped with TBP because it propagates track queries, despite using an external detector. The MOTIP rows use the published CVPR 2025 version, and SambaMOTR uses the published Deformable-DETR variant. Training stages and optimization schedules follow each method's architecture and learning objective; inference efficiency is measured under common conditions specified below.

\paragraph{Results on DanceTrack}

\begin{table}[htbp!]
  \centering
  \begingroup

  \caption{Comparison with state-of-the-art methods on the \textbf{DanceTrack} test set. Best and second-best results among Tracking-by-Propagation (TBP) methods are marked in \textbf{bold} and \underline{underlined}, respectively. Tracking-by-Detection-like methods are listed for reference only. \textsuperscript{‡} denotes additional data for tracking-model training, and \textsuperscript{†} denotes additional models. Train+val results are excluded from ranking.}
  \label{tab:dancetrack_sota}
  \setlength{\tabcolsep}{3pt}
  \begin{tabularx}{\textwidth}{@{}>{\raggedright\arraybackslash}Xccccc@{}}
    \toprule
    Method                                                                               & HOTA $\uparrow$  & DetA $\uparrow$  & AssA $\uparrow$  & MOTA $\uparrow$  & IDF1 $\uparrow$  \\
    \midrule
    \multicolumn{6}{l}{\textbf{Tracking-by-Detection-like}}                                                                                                                             \\
    Hybrid-SORT-ReID\textsuperscript{†}~\cite{yang2024hybridsort} (AAAI 2024)            & 65.7             & --               & --               & 91.8             & 67.4             \\
    DiffMOT\textsuperscript{†}~\cite{lv2024diffmot} (CVPR 2024)                          & 62.3             & 82.5             & 47.2             & 92.8             & 63.0             \\
    PlugTrack\textsuperscript{†}~\cite{kim2026plugtrack} (AAAI 2026)                     & 63.3             & 82.5             & 48.4             & 92.4             & 64.1             \\
    MOTIP~\cite{gao2025multiple} (CVPR 2025)                                             & 69.6             & 80.4             & 60.4             & 90.6             & 74.7             \\
    OA-SORT\textsuperscript{†}~\cite{li2026occlusion} (CVPR 2026)                        & 63.1             & --               & 48.5             & 91.7             & 64.2             \\
    \midrule
    \multicolumn{6}{l}{\textbf{Tracking-by-Propagation (TBP)}}                                                                                                                          \\
    MOTR~\cite{zeng2022motr} (ECCV 2022)                                                 & 54.2             & 73.5             & 40.2             & 79.7             & 51.5             \\
    MOTRv2\textsuperscript{†‡}~\cite{zhang2023motrv2} (CVPR 2023)                        & \underline{69.9} & \textbf{83.0}    & 59.0             & \textbf{91.9}    & 71.7             \\
    \textit{MeMOTR~\cite{gao2023memotr}} (ICCV 2023)                                     & 68.5             & 80.5             & 58.4             & 89.9             & 71.2             \\
    SambaMOTR~\cite{segu2025samba} (ICLR 2025)                                           & 67.2             & 78.8             & 57.5             & 88.1             & 70.5             \\
    TBDQ-Net\textsuperscript{†}~\cite{jia2026tbdq} (PR 2026)                             & 69.4             & 80.8             & \underline{59.8} & 89.5             & \underline{73.4} \\
    CO-MOT\textsuperscript{‡}~\cite{luo2025co} (ICLR 2025)                               & 69.4             & 82.1             & 58.9             & 91.2             & 71.9             \\
    TGFormer~\cite{zeng2025tgformer} (AAAI 2025)                                         & 69.1             & 81.9             & 58.3             & 91.3             & 71.9             \\
    DecoderTracker\textsuperscript{‡}~\cite{liao2026decodertracker} (PR 2026, train+val) & 60.8             & 81.2             & 47.1             & 91.1             & 62.9             \\
    \rowcolor{oursrow}
    \textbf{\textit{TAQ-MOTR (Ours)}}                                                    & \textbf{72.0}    & \underline{82.4} & \textbf{63.1}    & \underline{91.7} & \textbf{76.6}    \\
    \bottomrule
  \end{tabularx}
  \endgroup
\end{table}

In Table~\ref{tab:dancetrack_sota}, TAQ-MOTR achieves 72.0 HOTA, 82.4 DetA, 63.1 AssA, 91.7 MOTA, and 76.6 IDF1, improving over MeMOTR by 3.5, 1.9, 4.7, 1.8, and 5.4 points, respectively. It also exceeds CO-MOT by 2.6 HOTA and 4.2 AssA, supporting the benefit of tracking-aware query formation alongside auxiliary supervision. Compared with MOTRv2, gains of 2.1 HOTA, 4.1 AssA, and 4.9 IDF1 accompany slightly lower DetA and MOTA, indicating stronger association with competitive detection performance.

MOTRv2 and CO-MOT use CrowdHuman in addition to DanceTrack train; the listed DecoderTracker variant also uses DanceTrack val and is excluded from ranking. MOTRv2 obtains query priors from an external YOLOX detector, while TAQ-MOTR uses only DanceTrack train for tracker training without an additional detector. We report MOTRv2's 69.9-HOTA setting; its 73.4-HOTA variant additionally uses association post-processing, train+val training, and test-time ensembling~\cite{zhang2023motrv2}.

\paragraph{Results on SportsMOT}

\begin{table}[htbp!]
  \centering
  \begingroup

  \caption{Comparison with state-of-the-art methods on the \textbf{SportsMOT} test set. Best and second-best results among Tracking-by-Propagation (TBP) methods are marked in \textbf{bold} and \underline{underlined}, respectively. Tracking-by-Detection-like methods are listed for reference only. \textsuperscript{‡} denotes additional data for tracking-model training, and \textsuperscript{†} denotes additional models. Our mixed variant uses additional CrowdHuman data and is excluded from ranking.}
  \label{tab:sportsmot_sota}
  \setlength{\tabcolsep}{3pt}
  \begin{tabularx}{\textwidth}{@{}>{\raggedright\arraybackslash}Xccccc@{}}
    \toprule
    Method                                                       & HOTA $\uparrow$  & DetA $\uparrow$  & AssA $\uparrow$  & MOTA $\uparrow$  & IDF1 $\uparrow$  \\
    \midrule
    \multicolumn{6}{l}{\textbf{Tracking-by-Detection-like}}                                                                                                     \\
    DiffMOT\textsuperscript{†}~\cite{lv2024diffmot} (CVPR 2024)  & 72.1             & 86.0             & 60.5             & 94.5             & 72.8             \\
    MOTIP~\cite{gao2025multiple} (CVPR 2025)                     & 72.6             & 83.5             & 63.2             & 92.4             & 77.1             \\
    ETTrack\textsuperscript{†}~\cite{han2025ettrack} (AI 2025)   & 72.2             & 86.9             & 60.1             & 94.9             & 72.5             \\
    FDTA\textsuperscript{†}~\cite{shao2026detection} (CVPR 2026) & 74.2             & 84.1             & 65.5             & 93.0             & 78.5             \\
    \midrule
    \multicolumn{6}{l}{\textbf{Tracking-by-Propagation (TBP)}}                                                                                                  \\
    \textit{MeMOTR~\cite{gao2023memotr}} (ICCV 2023)             & 70.0             & \textbf{83.1}    & 59.1             & \underline{91.5} & 71.4             \\
    SambaMOTR~\cite{segu2025samba} (ICLR 2025)                   & 69.8             & 82.2             & 59.4             & 90.3             & 71.9             \\
    TBDQ-Net\textsuperscript{†}~\cite{jia2026tbdq} (PR 2026)     & \underline{72.1} & \underline{82.8} & \underline{62.8} & 89.5             & \underline{72.9} \\
    \rowcolor{oursrow}
    \textbf{\textit{TAQ-MOTR (Ours, train)}}                     & \textbf{72.3}    & 82.3             & \textbf{63.6}    & \textbf{92.0}    & \textbf{75.6}    \\
    \rowcolor{oursrow}
    \textit{TAQ-MOTR\textsuperscript{‡} (Ours, mixed)}           & 74.6             & 84.9             & 65.6             & 94.7             & 77.8             \\
    \bottomrule
  \end{tabularx}
  \endgroup
\end{table}

As shown in Table~\ref{tab:sportsmot_sota}, TAQ-MOTR achieves 72.3 HOTA, 82.3 DetA, 63.6 AssA, 92.0 MOTA, and 75.6 IDF1. Compared with MeMOTR, it improves HOTA, AssA, and IDF1 by 2.3, 4.5, and 4.2 points, respectively, while DetA is 0.8 points lower. These results indicate stronger identity association under rapid motion and dense interactions, with room to improve detection.

DiffMOT, ETTrack, and TBDQ-Net use YOLOX detections or proposal priors; FDTA uses Video Depth Anything for pseudo-depth supervision during training. Our main result uses only SportsMOT train and no additional model for detection or supervision. For reference, we also report a variant trained with additional CrowdHuman data, reaching 74.6 HOTA and 77.8 IDF1. This variant is excluded from the main ranking.

\paragraph{Results on BFT}

To align our benchmark coverage with recent methods such as MOTIP~\cite{gao2025multiple} and SambaMOTR~\cite{segu2025samba}, we additionally evaluate TAQ-MOTR on BFT. Table~\ref{tab:bft_sota} shows that TAQ-MOTR achieves the best results on all five metrics among the compared TBP methods, exceeding SambaMOTR by 3.8 HOTA, 4.7 AssA, and 3.7 IDF1. These gains indicate stronger association and identity preservation for non-human targets with complex motion. The SORT, ByteTrack, OC-SORT, and TrackFormer results are reported by NetTrack~\cite{zheng2024nettrack}; the first three use YOLOX detections. FDTA uses additional depth supervision.

\begin{table}[htbp!]
  \centering
  \begingroup

  \caption{Comparison with state-of-the-art methods on the \textbf{BFT} test set. Best and second-best results among Tracking-by-Propagation (TBP) methods are marked in \textbf{bold} and \underline{underlined}, respectively. Tracking-by-Detection-like methods are listed for reference only. \textsuperscript{‡} denotes additional data for tracking-model training, and \textsuperscript{†} denotes additional models.}
  \label{tab:bft_sota}
  \setlength{\tabcolsep}{3pt}
  \begin{tabularx}{\textwidth}{@{}>{\raggedright\arraybackslash}Xccccc@{}}
    \toprule
    Method                                                             & HOTA $\uparrow$  & DetA $\uparrow$  & AssA $\uparrow$  & MOTA $\uparrow$  & IDF1 $\uparrow$  \\
    \midrule
    \multicolumn{6}{l}{\textbf{Tracking-by-Detection-like}}                                                                                                           \\
    SORT\textsuperscript{†}~\cite{bewley2016sort} (ICIP 2016)          & 61.2             & 60.6             & 62.3             & 75.5             & 77.2             \\
    ByteTrack\textsuperscript{†}~\cite{zhang2022bytetrack} (ECCV 2022) & 62.5             & 61.2             & 64.1             & 77.2             & 82.3             \\
    OC-SORT\textsuperscript{†}~\cite{cao2022ocsort} (CVPR 2023)        & 66.8             & 65.4             & 68.7             & 77.1             & 79.3             \\
    MOTIP~\cite{gao2025multiple} (CVPR 2025)                           & 70.5             & 69.6             & 71.8             & 77.1             & 82.1             \\
    FDTA\textsuperscript{†}~\cite{shao2026detection} (CVPR 2026)       & 72.2             & 70.1             & 74.5             & 78.2             & 84.2             \\
    \midrule
    \multicolumn{6}{l}{\textbf{Tracking-by-Propagation (TBP)}}                                                                                                        \\
    TrackFormer~\cite{meinhardt2022trackformer} (CVPR 2022)            & 63.3             & \underline{66.0} & 61.1             & \underline{74.1} & 72.4             \\
    SambaMOTR~\cite{segu2025samba} (ICLR 2025)                         & \underline{69.6} & \underline{66.0} & \underline{73.6} & 72.0             & \underline{81.9} \\
    \rowcolor{oursrow}
    \textbf{\textit{TAQ-MOTR (Ours)}}                                  & \textbf{73.4}    & \textbf{69.0}    & \textbf{78.3}    & \textbf{76.7}    & \textbf{85.6}    \\
    \bottomrule
  \end{tabularx}
  \endgroup
\end{table}

\paragraph{Efficiency Analysis}

Efficiency is measured using the open-source implementations on a single NVIDIA GeForce RTX 4090 GPU, with sequential frame-by-frame inference and batch size 1. All methods use the same input resizing protocol: a target short side of 768 pixels and a maximum long side of 1536 pixels. FPS measures model inference time only, excluding preprocessing and post-processing. MOTRv2's reported 16.1 FPS includes both YOLOX and tracker inference.

\begin{table}[htbp!]
  \centering
  \begingroup
  \caption{Efficiency comparison with open-source TBP methods on DanceTrack. All methods follow the same measurement protocol. MOTRv2 includes YOLOX in both its parameter count and inference time.}
  \label{tab:efficiency_comparison}
  \begin{tabular}{lcc}
    \toprule
    Method                                     & Parameters (M)$\downarrow$ & FPS$\uparrow$ \\
    \midrule
    MOTR~\cite{zeng2022motr}                   & 43.9                       & 21.0          \\
    MOTRv2~\cite{zhang2023motrv2} (with YOLOX) & 140.6                      & 16.1          \\
    CO-MOT~\cite{luo2025co}                    & \textbf{41.7}              & 25.1          \\
    MeMOTR~\cite{gao2023memotr}                & 42.9                       & 22.5          \\
    \rowcolor{oursrow}
    \textbf{TAQ-MOTR}                          & 48.9                       & \textbf{41.1} \\
    \bottomrule
  \end{tabular}
  \endgroup
\end{table}

As shown in Table~\ref{tab:efficiency_comparison}, TAQ-MOTR reaches 41.1 FPS with 48.9M parameters. It runs faster than the compared TBP implementations. Its parameter count is slightly higher than those of MOTR, CO-MOT, and MeMOTR, but lower than that of MOTRv2 with YOLOX. The compact hybrid encoder contributes to this efficiency.

\begin{figure}[htbp!]
  \centering
  \includegraphics[width=\linewidth]{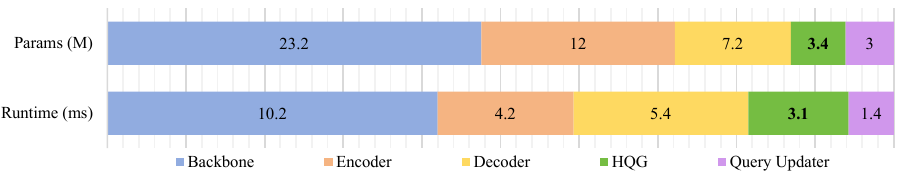}
  \caption{
    \textbf{Module-level efficiency analysis of TAQ-MOTR.}
    Rounded parameter counts and reported per-frame runtimes are shown for the backbone, hybrid encoder, decoder, HQG, and query updater.}
  \label{fig:efficiency_analysis}
\end{figure}

Fig.~\ref{fig:efficiency_analysis} shows that HQG contains 3.4M parameters (approximately 7.0\% of TAQ-MOTR's 48.9M parameters) and takes 3.1 ms per frame.
Most computation remains in feature extraction and decoder processing, indicating that tracking-aware query formation adds limited overhead.

\subsection{Ablation Study and Analysis}

We conduct ablation studies on the DanceTrack validation set to validate the main design choices of TAQ-MOTR. HOTA is used as the primary overall metric, while DetA, AssA, and IDF1 are reported to analyze detection accuracy, association quality, and identity preservation, respectively. Unless otherwise specified, all variants use the same training schedule and inference protocol as the default model, and the best result of each metric is shown in bold.

\paragraph{Prior carriers in HQG}
\begin{table}[htbp!]
  \centering
  \caption{Ablation of HQG priors on the full DanceTrack validation set. DP, TM, and LM denote detection proposals, track memory, and learnable memory, respectively. Original reports the published MeMOTR results~\cite{gao2023memotr}. Unified basic model adopts the same backbone, encoder, and pretrained weights as TAQ-MOTR while retaining the rest of MeMOTR's tracking design. The best result for each metric is shown in bold, and unreported values are indicated by a dash.}
  \label{tab:ablation_hqg_priors}
  \setlength{\tabcolsep}{2pt}
  \begin{tabular}{@{}ccccccccc@{}}
    \toprule
    DP                                      & TM         & LM         & HOTA $\uparrow$ & DetA $\uparrow$ & AssA $\uparrow$ & IDF1 $\uparrow$ & DetRe $\uparrow$ & IDSW $\downarrow$ \\
    \midrule
    \multicolumn{9}{l}{\textbf{Baselines}}                                                                                                                                           \\
    \multicolumn{3}{c}{Original}            & 63.9       & 74.6       & 55.0            & 67.1            & --              & 1237                                                   \\
    \multicolumn{3}{c}{Unified basic model} & 62.9       & 76.0       & 52.3            & 66.1            & 85.4            & 1530                                                   \\
    \midrule
    \multicolumn{9}{l}{\textbf{HQG variants}}                                                                                                                                        \\
    \Checkmark                              & --         & --         & 58.1            & 71.9            & 47.1            & 61.9            & --               & --                \\
    \Checkmark                              & \Checkmark & --         & 63.3            & 74.1            & 54.2            & 68.0            & 83.7             & \textbf{996}      \\
    \Checkmark                              & --         & \Checkmark & 64.4            & \textbf{77.1}   & 53.9            & 69.1            & --               & --                \\
    \rowcolor{oursrow}
    \Checkmark                              & \Checkmark & \Checkmark & \textbf{65.3}   & 76.5            & \textbf{56.0}   & \textbf{70.3}   & \textbf{86.1}    & 1038              \\
    \bottomrule
  \end{tabular}
\end{table}

Relative to DP+LM, introducing track memory improves HOTA, AssA, and IDF1 by 0.9, 2.1, and 1.2 points, respectively, with a 0.6-point decrease in DetA. These results highlight the detection--association tradeoff introduced by track-state conditioning.

In the Baselines section of Table~\ref{tab:ablation_hqg_priors}, Original reports the published MeMOTR results~\cite{gao2023memotr}, whereas Unified basic model reports the results of a MeMOTR control using the same backbone, encoder, and pretrained weights as TAQ-MOTR. The underlying feature-extraction architecture and pretraining are aligned, with corresponding adaptations to the training settings for the MeMOTR baseline. The comparison therefore provides an additional reference for assessing HQG rather than fully isolating its contribution. Compared with Original, Unified basic model improves DetA from 74.6 to 76.0 but reduces AssA from 55.0 to 52.3 and IDF1 from 67.1 to 66.1. As discussed in FDTA~\cite{shao2026detection}, object embeddings optimized solely for detection tend to have weaker discriminative ability for association. In our setting, the backbone and encoder initialized with RT-DETRv2 detection weights may favor detection cues, potentially contributing to the observed improvement in DetA and decline in association.

We additionally report detection recall (DetRe) and identity switches (IDSW) to assess missed detections and identity consistency. Relative to Unified basic model, the full HQG configuration improves HOTA, DetA, AssA, and IDF1 by 2.4, 0.5, 3.7, and 4.2 points, respectively. It also increases DetRe from 85.4 to 86.1 and reduces IDSW from 1,530 to 1,038. These results support improved overall target recall and identity association, without separately establishing a gain in recall for currently untracked objects. However, compared with DP+TM, the full configuration improves DetRe and IDF1 while increasing IDSW from 996 to 1,038, indicating a tradeoff between recall and identity switches.

\paragraph{Image-based detection proposals}
\begin{table}[htbp!]
  \centering
  \caption{Ablation of feature levels for constructing image-based detection proposals on the DanceTrack validation set.}
  \label{tab:ablation_feature_levels}
  \setlength{\tabcolsep}{4pt}
  \begin{tabular}{lcccc}
    \toprule
    Level            & HOTA $\uparrow$ & DetA $\uparrow$ & AssA $\uparrow$ & IDF1 $\uparrow$ \\
    \midrule
    Stride 8         & 59.5            & 66.7            & 53.2            & 68.2            \\
    Stride 16        & 61.2            & 68.6            & 54.7            & 68.9            \\
    Stride 16+32     & 64.8            & 74.7            & \textbf{56.4}   & 70.0            \\
    \rowcolor{oursrow}
    Stride 32 (Ours) & \textbf{65.3}   & \textbf{76.5}   & 56.0            & \textbf{70.3}   \\
    \bottomrule
  \end{tabular}
\end{table}

Table~\ref{tab:ablation_feature_levels} compares proposal feature levels. The stride-32 setting has the highest reported HOTA, DetA, and IDF1; stride-16+32 has the highest AssA but requires additional computation. We use the compact stride-32 representation as the default. The differences suggest that denser proposal features do not necessarily improve overall tracking.

\paragraph{Track memory}
\begin{table}[htbp!]
  \centering
  \caption{Ablation of retaining recently lost tracks in HQG track memory.}
  \label{tab:ablation_lost_track_memory}
  \setlength{\tabcolsep}{5pt}
  \begin{tabular}{lcccc}
    \toprule
    \makecell{Retain lost\\tracks} & HOTA $\uparrow$ & DetA $\uparrow$ & AssA $\uparrow$ & IDF1 $\uparrow$ \\
    \midrule
    \rowcolor{oursrow}
    0 frames (Ours)                & \textbf{65.3}   & 76.5            & \textbf{56.0}   & \textbf{70.3}   \\
    $\leq$1 frame                  & 61.7            & \textbf{78.3}   & 48.8            & 63.1            \\
    $\leq$3 frames                 & 63.9            & 77.9            & 52.5            & 66.3            \\
    $\leq$5 frames                 & 63.7            & 77.5            & 52.5            & 66.4            \\
    \bottomrule
  \end{tabular}
\end{table}

Table~\ref{tab:ablation_lost_track_memory} compares active-only HQG memory with memory that also retains recently lost tracks. The active-only default gives the highest HOTA, AssA, and IDF1, while including lost tracks improves DetA. Temporarily lost queries still remain in the decoder and retain their identities. These results favor active track states for HQG conditioning; the additional historical cues may aid detection while reducing association quality.

\paragraph{Learnable memory}
\begin{table}[htbp!]
  \centering
  \caption{Effect of memory type and shared-set token count in HQG on the DanceTrack validation set. All HQG layers use the same memory set. Fixed random memory uses frozen, randomly initialized embeddings.}
  \label{tab:hqg_learnable_memory}
  \setlength{\tabcolsep}{5pt}
  \begin{tabular}{ccccc}
    \toprule
    Tokens (shared set)  & HOTA $\uparrow$ & DetA $\uparrow$ & AssA $\uparrow$ & IDF1 $\uparrow$ \\
    \midrule
    0 (None)             & 63.3            & 74.1            & 54.2            & 68.0            \\
    4 (Learnable)        & 64.2            & 74.6            & 55.3            & 70.2            \\
    32 (Fixed random)    & 64.6            & 74.7            & 56.0            & 69.6            \\
    \rowcolor{oursrow}
    32 (Learnable, Ours) & 65.3            & \textbf{76.5}   & 56.0            & 70.3            \\
    64 (Learnable)       & \textbf{65.7}   & 76.2            & \textbf{56.6}   & \textbf{70.8}   \\
    \bottomrule
  \end{tabular}
\end{table}

Inspired by the learnable detection queries in \mbox{MeMOTR}~\cite{gao2023memotr}, HQG uses a shared set of trainable memory tokens across all layers to encode dataset-level priors. Table~\ref{tab:hqg_learnable_memory} shows that using a shared set of 32 tokens improves HOTA from 63.3 to 65.3. Increasing the count to 64 yields further gains of 0.4 HOTA, 0.6 AssA, and 0.5 IDF1, but reduces DetA by 0.3. We select 32 as an efficiency/performance compromise, retaining competitive accuracy with fewer tokens participating in cross-attention.

To examine memory learnability, we add a fixed random memory control with the same shared set of 32 token slots, embedding dimensions, and attention operations as the learnable variant. Compared with this control, learnable memory achieves gains of 0.7 HOTA, 1.8 DetA, and 0.7 IDF1, while both obtain 56.0 AssA. These results support the contribution of learned detection priors. Freezing the embeddings reduces the number of trainable parameters while keeping the memory representation size unchanged. Fixed memory also improves HOTA and IDF1 by 1.3 and 1.6 points over using no memory, suggesting that the network can learn to use these tokens even when their embeddings are fixed. Learning the embeddings further allows the memory itself to adapt to the dataset. More effective memory representations and learning strategies remain directions for future work.

\paragraph{Number of HQG layers}
\begin{table}[htbp!]
  \centering
  \caption{Ablation of the number of HQG layers.}
  \label{tab:ablation_hqg_depth}
  \setlength{\tabcolsep}{5pt}
  \begin{tabular}{ccccc}
    \toprule
    Layers   & HOTA $\uparrow$ & DetA $\uparrow$ & AssA $\uparrow$ & IDF1 $\uparrow$ \\
    \midrule
    1        & 63.7            & \textbf{76.8}   & 52.9            & 68.4            \\
    2        & 64.1            & 76.1            & 54.3            & 69.0            \\
    \rowcolor{oursrow}
    3 (Ours) & \textbf{65.3}   & 76.5            & \textbf{56.0}   & \textbf{70.3}   \\
    6        & 63.7            & 75.2            & 54.1            & 68.2            \\
    \bottomrule
  \end{tabular}
\end{table}

Table~\ref{tab:ablation_hqg_depth} compares HQG depth. Three layers yield the highest HOTA, AssA, and IDF1, whereas one layer gives the highest DetA. Six layers reduce overall performance. We therefore use three layers as the default.

\paragraph{Detection-only auxiliary learning}
\begin{table}[htbp!]
  \centering
  \caption{Auxiliary supervision schedules on the DanceTrack validation set. Parenthesized epochs indicate when intermediate-layer auxiliary decoder losses are removed and when the entire auxiliary branch is disabled, respectively. None denotes training without auxiliary supervision.}
  \label{tab:ablation_det_aux_schedule}
  \setlength{\tabcolsep}{5pt}
  \begin{tabular}{lcccc}
    \toprule
    Auxiliary schedule     & HOTA $\uparrow$ & DetA $\uparrow$ & AssA $\uparrow$ & IDF1 $\uparrow$ \\
    \midrule
    None (0, 0)            & 64.2            & 74.8            & 55.2            & 69.3            \\
    Staged (8, 24)         & 64.6            & 75.3            & 55.7            & 69.0            \\
    \rowcolor{oursrow}
    Staged (14, 24) (Ours) & \textbf{65.3}   & 76.5            & \textbf{56.0}   & \textbf{70.3}   \\
    Staged (21, 24)        & 64.8            & 76.7            & 55.6            & 69.1            \\
    Staged (24, 24)        & 65.0            & \textbf{77.0}   & 54.9            & 68.6            \\
    Staged (14, 21)        & 64.4            & 75.2            & 55.3            & 70.1            \\
    Staged (14, 27)        & 64.9            & 76.7            & 55.4            & 67.7            \\
    \bottomrule
  \end{tabular}
\end{table}

The supervision imbalance in Fig.~\ref{fig:label_assignments} motivates additional detection supervision without changing the main branch's association label assignment. Table~\ref{tab:ablation_det_aux_schedule} shows that our default auxiliary schedule improves HOTA, DetA, AssA, and IDF1 over training without the branch by 1.1, 1.7, 0.8, and 1.0 points, respectively. The timing of intermediate-layer loss removal and complete auxiliary-branch shutdown also affects the detection--association balance. Relative to the default, removing intermediate-layer losses earlier reduces DetA by 1.2 points, whereas delaying their removal increases DetA by 0.2--0.5 points but lowers AssA by 0.4--1.1 points. Advancing complete branch shutdown reduces DetA by 1.3 points; postponing it yields only a 0.2-point DetA gain while lowering IDF1 by 2.6 points. These results support sufficient auxiliary supervision during early and middle training, followed by its reduction and removal to focus late optimization on the main MOT objective.

\paragraph{Tracked-object complementation}
We compare models with and without HQG on DanceTrack validation, grouping detection queries after the first detection-only decoder layer.
A detection query is classified as related if its box IoU with the current-frame GT box corresponding to the track query's assigned ID is $\geq 0.5$, and as unrelated if it is $\leq 0.1$.
We then use self-attention masks to control information flow from detection queries to track queries and test whether target-related information improves tracking.
A masks unrelated queries matched to the related set in number and approximately in classification score; B masks the related set; C uses normal self-attention.
Masks block only track-query access to these sets in all joint self-attention layers; tracks with insufficient controls remain unmasked.
Each setting propagates its own trajectories, with GT used only to define the diagnostic masks.

\begin{table}[H]
  \centering
  \caption{\textbf{Effect of target-related detection-to-track interaction on DanceTrack.}
    A: unrelated masking; B: related masking; C: normal self-attention.
    The C rows are reference results for the corresponding models, evaluated with different inference settings from A and B.
    All metrics are percentages.
    Signed changes denote absolute differences on the percentage scale.}
  \label{tab:sa_masking_complementation}
  \setlength{\tabcolsep}{4pt}
  \begin{tabular}{@{}lcccc@{}}
    \toprule
    Configuration & HOTA $\uparrow$ & DetA $\uparrow$ & AssA $\uparrow$ & IDF1 $\uparrow$ \\
    \midrule
    w/o HQG + A   & 62.7            & 76.1            & 51.9            & 66.1            \\
    w/o HQG + B   & 60.0            & 75.7            & 47.7            & 63.0            \\
    w/o HQG + C   & 62.9            & 76.0            & 52.3            & 66.1            \\
    w/ HQG + A    & 65.1            & 76.3            & 55.6            & 69.7            \\
    w/ HQG + B    & 57.0            & 74.9            & 43.5            & 60.9            \\
    \rowcolor{oursrow}
    w/ HQG + C    & \textbf{65.3}   & \textbf{76.5}   & \textbf{56.0}   & \textbf{70.3}   \\
    \bottomrule
  \end{tabular}
\end{table}

As shown in Table~\ref{tab:sa_masking_complementation}, related masking (B) decreases HOTA relative to matched unrelated masking (A), with changes of $-8.1\%$ and $-2.7\%$ for the models with and without HQG, respectively.
The larger performance decrease with HQG suggests that information from its target-related detection queries makes a greater contribution to tracking performance.
For the model with HQG, AssA and IDF1 also decrease from A to B ($-12.1\%$ and $-8.8\%$, respectively), indicating that this information contributes to association accuracy and identity preservation.
These results indicate that track queries can obtain complementary information from HQG-selected detection queries through self-attention to improve tracking performance.

\subsection{Visual Analysis}

\paragraph{Tracking Results Visualization}
\begin{figure*}[ht!]
  \centering
  \includegraphics[width=\textwidth]{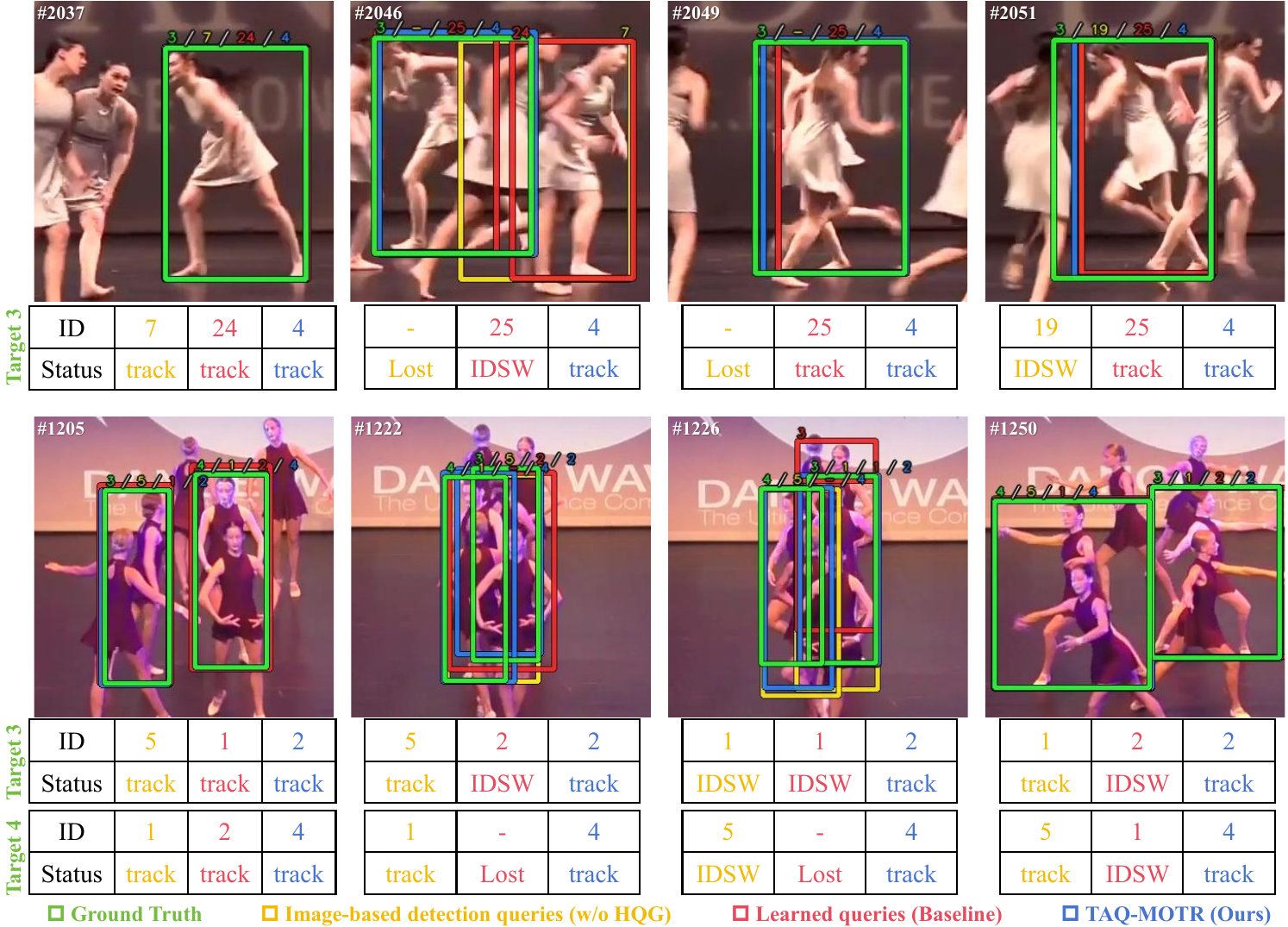}
  \caption{
    \textbf{Qualitative comparison of detection-query formation strategies on DanceTrack.}
    Green boxes denote ground truth. Yellow, red, and blue boxes denote tracking outputs from three query-formation variants: image-based detection queries without HQG, learned queries, and HQG-generated queries (TAQ-MOTR), respectively.
    The status tables report matched IDs and tracking states: ``Lost'' denotes no matched result, ``IDSW'' an identity switch, and ``track'' correct tracking.
    In the illustrated frames, TAQ-MOTR preserves the queried identities through occlusion, crossing, and overlap; this comparison provides qualitative evidence of association behavior.}
  \label{fig:query_box_visualization}
\end{figure*}

Fig.~\ref{fig:query_box_visualization} compares detection-query formation strategies on crowded DanceTrack clips with occlusion, crossing, and target overlap. These conditions challenge propagated track queries and motivate complementary current-frame evidence. Image-based detection queries use RT-DETR-style selection from current-frame encoder features without tracking-aware interaction.

In the first row, heavy occlusion around frame~\#2046 causes image-based detection queries to lose the target, while learned queries assign a different identity. Image-based detection queries subsequently recover the more visible target with an inconsistent ID. This sequence highlights the difficulty of maintaining identity from current-frame visual evidence alone when the local region remains ambiguous. The second row shows identity switches between interacting targets, with learned queries losing Target~4 during crossing. TAQ-MOTR maintains the queried targets in the tracked state across these frames. These examples illustrate more stable association and are consistent with the intended role of tracking-aware query formation, but do not isolate the mechanism responsible for the improvement.

\paragraph{Query Focus Visualization}
\begin{figure*}[htbp!]
  \centering
  \includegraphics[width=\textwidth]{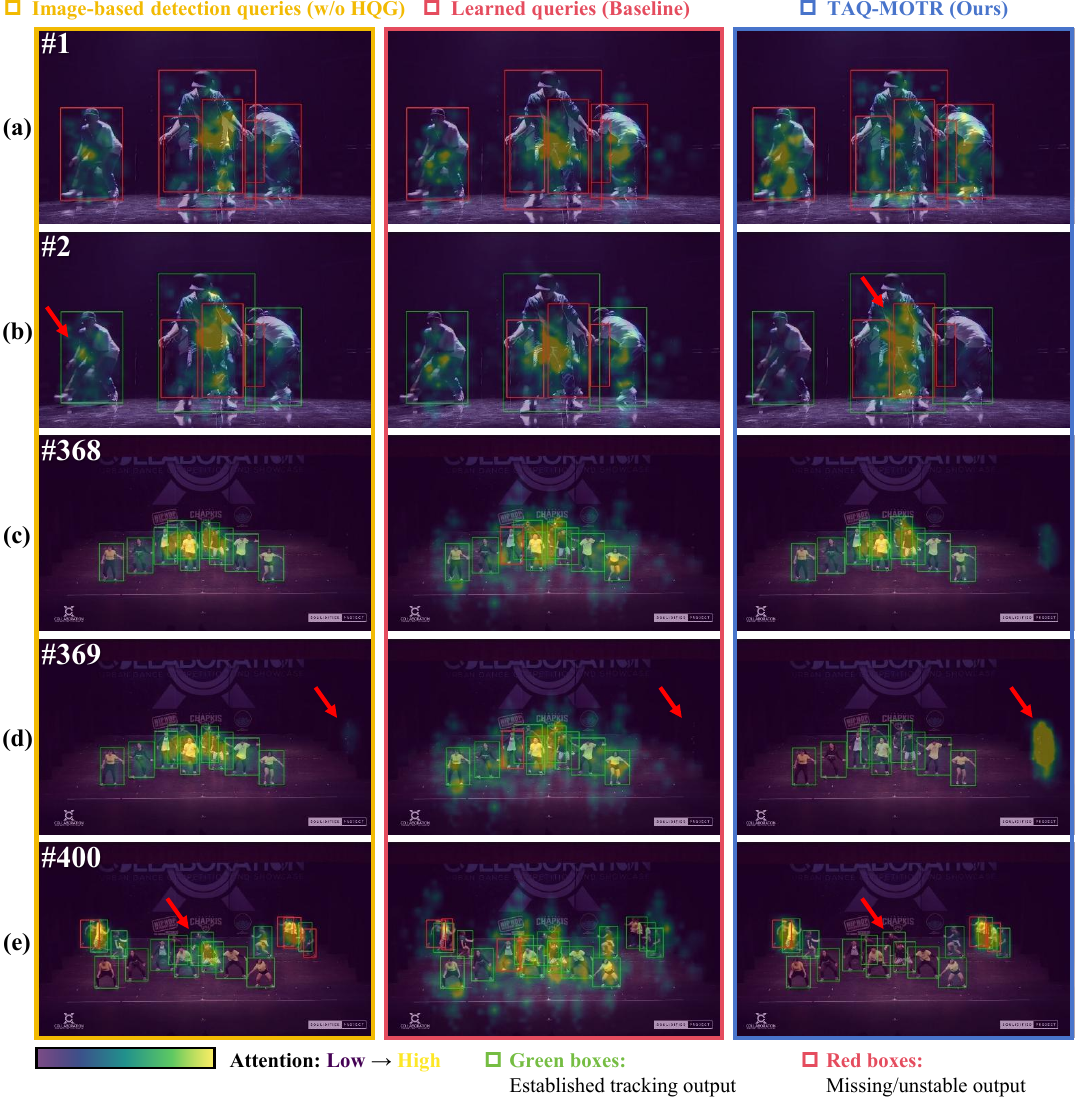}
  \caption{
    \textbf{Query-focus visualization of detection queries on DanceTrack.}
    The heatmaps aggregate cross-attention between detection queries and multi-scale image features in the first decoder layer; each panel is independently normalized, so heatmap colors indicate relative spatial focus rather than absolute attention magnitudes.
    Green boxes mark established tracking outputs and red boxes missing or unstable outputs; these colors do not indicate track-query retention. Panels (a)--(b) show video initialization, and (c)--(e) illustrate changes in query focus in a crowded scene.}
  \label{fig:query_focus_visualization}
\end{figure*}

In Fig.~\ref{fig:query_focus_visualization}(a)--(b), all three variants attend to targets during initialization, but TAQ-MOTR shifts its focus toward the overlapping central targets as track states become available.
Learned queries show a more diffuse pattern in the crowded scene of Fig.~\ref{fig:query_focus_visualization}(c)--(e).
In (d), image-based and learned queries continue to emphasize the central group, whereas TAQ-MOTR shifts toward the rightmost target indicated by the arrow.
TAQ-MOTR then redistributes its attention across other target regions in (e).
These patterns are consistent with detection-query focus adapting to current-frame evidence and existing track states.

\begin{figure}[htbp!]
  \centering
  \includegraphics[width=\linewidth]{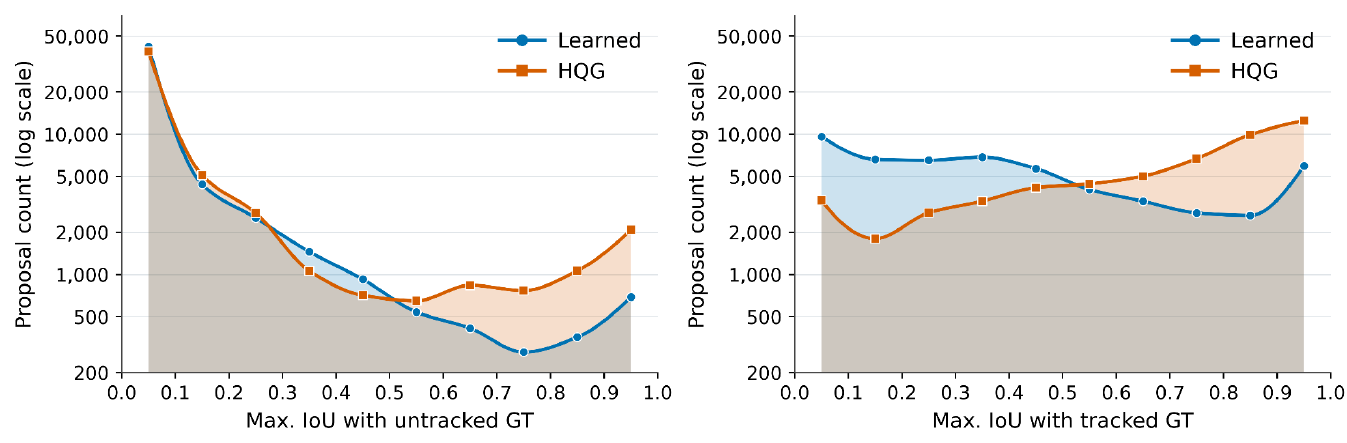}
  \caption{\textbf{IoU distributions of learned and HQG-initialized detection queries on DanceTrack.}
    The horizontal axis shows each query box's maximum IoU with currently untracked (left) or tracked (right) GT; the logarithmic vertical axis shows query counts.}
  \label{fig:query_gt_alignment}
\end{figure}

Fig.~\ref{fig:query_gt_alignment} further examines the alignment of detection-query boxes from the first detection-only decoder layer with currently untracked and tracked objects.
Tracked GT identities are those bound to retained input track queries.
We compare the two methods on common DanceTrack validation frames containing both groups, with identical tracking-state assignments.
We compute each box's maximum IoU with each GT group separately, yielding query-level distributions.
Compared with learned queries, HQG-initialized queries not only yield more high-IoU boxes for currently untracked objects but also concentrate at higher IoU for tracked objects, with the latter distribution peaking in the 0.9--1.0 interval.
These results indicate that HQG-initialized detection queries are more concentrated on object regions and achieve better spatial alignment than learned queries.

\section{Limitations}

HQG does not explicitly estimate or gate prior reliability. Under dense overlap and similar appearance, image proposals may become ambiguous and propagated tracks may drift or confuse identities. The lost-track ablation in Table~\ref{tab:ablation_lost_track_memory} shows that adding uncertain track states can improve detection but degrade association and identity preservation. Improving robustness to unreliable priors remains an open direction.

The compact stride-32 proposals in Table~\ref{tab:ablation_feature_levels} may also limit sensitivity to small targets, local overlaps, and abrupt shape changes. Selective use of finer multi-scale evidence could address this limitation while preserving efficient query formation.

Finally, auxiliary supervision relies on a coarse, manually calibrated schedule whose transition points influence the balance between detection and association performance. Learning to adapt the strength and duration of auxiliary supervision to the model's training progress remains an important direction for future work.

\section{Conclusion}
TAQ-MOTR forms detection queries from current-frame image evidence, track memory, and learnable priors before shared decoding. A training-only auxiliary branch strengthens detection learning through independent label assignments while preserving identity-based assignment in the main branch. Experiments on DanceTrack, SportsMOT, and BFT show improved association and identity preservation, with the highest HOTA among the compared TBP methods. Ablations support tracking-aware query formation and staged auxiliary supervision. The results motivate conditioning detection queries on evolving track states while improving the robustness of memory priors and auxiliary scheduling.

\section*{Declaration of generative AI and AI-assisted technologies in the manuscript preparation process}

During the preparation of this work, the authors used Codex (OpenAI) to refine the academic phrasing, improve the overall grammatical structure of the manuscript, ensure correct spelling and formatting, and assist in organizing the public code repository and its accompanying documentation. After using this tool, the authors reviewed and edited the content as needed and take full responsibility for the content of the published article.

\section*{CRediT authorship contribution statement}
Hanyang Liu: Writing – review \& editing, Writing – original draft, Software, Visualization, Validation, Formal analysis, Data curation, Conceptualization;
Hong Zhang: Resources, Methodology, Formal analysis, Funding acquisition;
Ding Yuan: Writing – review \& editing, Resources,  Formal analysis;
Yang Han: Data curation, Investigation;
Yifan Yang: Supervision, Project administration, Funding acquisition.

\section*{Data availability}
The source code, trained checkpoints, training logs, and tracking results are publicly available at \url{https://github.com/Magicmiya/TAQ-MOTR}.

\section*{Acknowledgements}
This work was supported by the National Natural Science Foundation of China (Grant Nos. 62433003 and 62476017).

\bibliographystyle{elsarticle-num}
\balance
\bibliography{main}

\end{document}